\documentclass[10pt,twocolumn,twoside]{IEEEtran}

\usepackage{graphics}
\usepackage[dvipsnames]{xcolor}
\usepackage{amsmath,amsfonts,amssymb}
\usepackage{pstricks,pstricks-add}
\usepackage{amsmath}
\usepackage{cite}

\usepackage{enumerate}
\usepackage[shortlabels]{enumitem}

\usepackage{algorithm2e}
\SetAlCapSkip{1em}

\usepackage{array}
\usepackage{epsfig} 
\usepackage{subcaption}
\usepackage{multicol}
\usepackage{multirow}

\makeatletter
\def\BState{\State\hskip-\ALG@thistlm}
\makeatother

\IEEEoverridecommandlockouts                              

\newtheorem{theorem}{Theorem}[section]

\newtheorem{defn}[theorem]{Definition}
\newtheorem{ex}[theorem]{Example}

\title{A Reinforcement Learning Framework for \\ Sequencing Multi-Robot Behaviors*}

\author{Pietro Pierpaoli, Thinh T. Doan, Justin Romberg, and Magnus Egerstedt $^{\dagger}$
\thanks{*This work was supported by the Army Research Lab under Grant DCIST CRA W911NF-17-2-0181.}
\thanks{$^{\dagger}$ The authors are with the School of Electrical and Computer Engineering, Georgia Institute of Technology, Atlanta, GA 30332, USA. {\tt\small \{pietro.pierpaoli,thinhdoan, magnus\}@gatech.edu, jrom@ece.gatech.edu}}} 
\begin{document}

\maketitle
\thispagestyle{empty}
\pagestyle{empty}

\begin{abstract}
Given a list of behaviors and associated parameterized controllers for solving different individual tasks, we study the problem of selecting an optimal sequence of coordinated behaviors in multi-robot systems for completing a given mission, which could not be handled by any single behavior. In addition, we are interested in the case where partial information of the underlying mission is unknown, therefore, the robots must cooperatively learn this information through their course of actions. Such problem can be formulated as an optimal decision problem in multi-robot systems, however, it is in general intractable due to modeling imperfections and the curse of dimensionality of the decision variables. To circumvent these issues, we first consider an alternate formulation of the original problem through introducing a sequence of behaviors' switching times. Our main contribution is then to propose a novel reinforcement learning based method, that combines Q-learning and online gradient descent, for solving this reformulated problem. In particular, the optimal sequence of the robots' behaviors is found by using Q-learning while the optimal parameters of the associated controllers are obtained through an online gradient descent method. Finally, to illustrate the effectiveness of our proposed method we implement it on a team of differential-drive robots for solving two different missions, namely, convoy protection and object manipulation.

\end{abstract}

\section{Introduction} \label{sec:introduction}

In multi-robot systems, there has been a great success in designing distributed controllers to address individual tasks through coordination \cite{oh2015survey,cortes2017coordinated,antonelli2013interconnected,schwager2011unifying}. However, many complex tasks in real-world applications require the robots being capable of performing beyond what has been achieved with task-oriented controllers. In addition, it is often that the information (or model) of the underlying tasks required by the existing controllers is partially known or imperfect, making solving these tasks nontrivial. For example, consider a team of robots tasked with moving a box between two points without previous knowledge of the box's physical properties (e.g. mass distribution, friction with the ground). Robots should first localize the box by exploring the space. Then, depending on their capabilities and available behaviors, the box should be pushed or lifted towards its destination.    

Through the paradigm of behavioral robotics~\cite{arkin1998behavior} it is possible to address this complexity by sequentially combining multi-robot task-oriented controllers (or {\it behaviors}); see for example \cite{nagavalli2017automated,pierpaoli2019sequential}. In addition, to achieve a long-term autonomy and an interaction with imperfect model the robots have to choose actions based on the real-time feedback observed from the systems. The idea of learning by interacting with the system (or environment) is the main theme in  reinforcement learning \cite{bertsekas2019reinforcement,Sutton2018_book}.

Our focus in this paper is to study the problem of selecting a sequence of coordinated behaviors and the associated parameterized controllers in  multi-robot systems for achieving a given task, which could not be handled by any individual behavior. In addition, we consider the applications where a partial information of the underlying task is unknown, therefore, the robots must cooperatively learn this information through their course of actions. Such problem can be formulated as an optimal decision problem in multi-robot systems. However, it is in general intractable due to the imperfection of the problem model and the curse of dimensionality of the decision variables. To address these issues, we first formulate a behaviors switching condition based on the dynamic of the behavior itself. Our main contribution is then to propose a novel reinforcement learning based method, a combination of Q-learning and online gradient descent to find the optimal sequence of the robots' behaviors and the optimal parameters of the associated controllers, respectively. Finally, to illustrate the effectiveness of our proposed method we implement it on a team of differential-drive robots for solving two different tasks, namely, convoy protection and simplified object manipulation.

Our paper is organized as follows. We start by describing a class of weighted-consensus coordinated behaviors in Section~\ref{sec:coordBeh}. The behavior selection problem is then formulated as in Section~\ref{sec:problemForm}, while our proposed method for solving this problem is presented in Section \ref{sec:qlearning}. We conclude this paper by illustrating two applications of the proposed technique to a multi-robot systems in Section~\ref{sec:applications}.

\subsection{Related Work}
A common approach in the design of multi-robot controllers for solving specific tasks is through the use of local control rules, whose collective implementation results in the desired behavior. Weighted consensus protocols can be employed to this end. See for example \cite{cortes2017coordinated} and references therein. The advantage of stating the multi-agent control problem in terms of task-specific controllers resides in the fact that provable guarantees exist on their convergence~\cite{zelazo2018graph}. 


The fundamental idea behind behavioral robotics \cite{arkin1998behavior} is to compose task-specific primitives into sequential or hierarchical controllers. This idea has been largely investigated for single robot systems and less so for multi-robot teams, where constrains on the flow of information prevents direct application of the single robots' algorithms. Proposed techniques for the composition of primitive controllers include formal methods~\cite{kress2018synthesis}, path planning~\cite{nagavalli2017automated}, Finite State Machines~\cite{marino2009behavioral}, Petri Nets~\cite{klavins2000formalism}, and Behavior Trees~\cite{colledanchise2014behavior}. Once an appropriate sequence of controllers is chosen, transitions between individual controllers must be feasible. Solutions to this problem include Motion Description Language~\cite{martin2012hybrid}, Graph Process Specifications~\cite{twu2010graph} and Control Barrier Functions~\cite{li2018formally,pierpaoli2019sequential}.


Finally, reinforcement learning offers a general paradigm for learning optimal policies in stochastic control problems based on simulation \cite{bertsekas2019reinforcement,Sutton2018_book}. In this context, an agent seeks to find an optimal policy through interacting with the unknown environment with the goal of optimizing its long-term future reward. Motivated by broad applications of the multi-agent systems, for example, mobile sensor networks \cite{CortesMKB2004,Ogren2004_TAC} and power networks \cite{Kar2013_QDLearning}, there is a growing interest in studying multi-agent reinforcement learning; see for example \cite{zhang2018networked,Wai2018_NIPS,DoanMR2019_DTD(0)} and the references therein. Our goal in this paper is to consider another interesting application of multi-agent reinforcement learning in solving the optimal behaviors selection problem over multi-robot systems.


\section{Coordinated Control of Multi-Robot Systems} \label{sec:coordBeh}
In this section, we provide some preliminaries for the main problem formulation. We start with some general definitions in multi-robot systems and then formally define the notion of behaviors used throughout the paper.    
\subsection{Multi-Robot Systems}
Consider a team of $N$ robots operating in a $2$-dimensional domain, where we denote by $x_{i}\in\mathbb{R}^2$ the state of robot $i$, for $i=1,\dots,N$. In addition, the dynamic of the robots is governed by a single integrator given as
\begin{equation} \label{eq:sing_int}
\dot{x}_{i} =  u_{i},
\end{equation}
where $u_{i}$ is the controller at robot $i$, which is a function of $x_i$ and the states of the robots interacting with robot $i$. The pattern of interactions between the robots is presented by an undirected graph $G = (V,E)$, where $V = \{1,\ldots,N\}$ and $E = (V\times V)$ are the index set and the set of pairwise interactions between the robots, respectively. Moreover, let $N_i = \{j\in V\,|\, (i,j)\in E\}$ be the neighboring set of robot $i$.


Each controller $u_i$ at robot $i$ is composed of two components, one only depends on its own state while the other represents the interaction with its neighbors. In particular, the controller $u_i: \mathbb{R}^{2+2|\mathcal{N}_i|} \mapsto \mathbb{R}^2$ in~(\ref{eq:sing_int}) is given as
\begin{equation} \label{eq:agent_controller}
u_i = -\sum_{j \in N_i} \left( \, w(x_i,x_j,\theta)(x_i - x_j) \, \right) + v(x_i,\phi), 
\end{equation}
where $w: \mathbb{R}^2\times \mathbb{R}^2 \times \Theta \rightarrow \mathbb{R}$, often referred to as an {\it edge weight function} \cite{mesbahi2010graph}, depends on the states of robot $i$ and its neighbors, and the parameter $\theta\in\Theta$. In addition, $v: \mathbb{R}^2 \times \Phi \rightarrow \mathbb{R}^2$ is the state-feedback term at the robot $i$, which depends only on its own state $x_i$ and a parameter $\phi\in\Phi$ representing robot $i$'s preference. Here, $\Theta$ and $\Phi$ are the feasible sets of the parameters $\theta$ and $\phi$, respectively. A concrete example of such controller together with the associated parameters will be given in the next section. Finally, as studied in \cite{cortes2017coordinated}, one can define an appropriate energy function  $\mathcal{E}: \mathbb{R}^{2N} \mapsto  \mathbb{R}_{\geq 0}$ with respect to the graph $G$, where the controller in \eqref{eq:agent_controller} can be described as the negativity of the local gradient of $\mathcal{E}$, i.e.,
\begin{equation} \label{eq:energy}
u_i = -\frac{\partial \mathcal{E}}{\partial x_i}\cdot
\end{equation}
This observation will be useful for our later development. 


\subsection{Coordinated Behaviors in Multi-Robot Systems} 
Given a collection of behaviors, our goal is to optimally select a sequence of them in order to complete a given mission. 


\begin{defn}\label{def:behavior}
A coordinated behavior $\mathcal{B}$ is defined by $5-$tuple
\begin{equation} \label{eq:behavior}
        \mathcal{B} = (w, \Theta, v, \Phi, G)
\end{equation}
where $\Theta$ and $\Phi$ are feasible sets for the parameters of controller \eqref{eq:agent_controller}. Moreover, $G$ is the graph representing the interaction structure between the robots. 

In addition, given $M$ distinct behaviors we compactly represent them as a library of behaviors $\mathcal{L}$ \begin{equation} \label{eq:libDef}
        \mathcal{L}_b=\{\mathcal{B}_1, \dots, \mathcal{B}_M\},
\end{equation}
where each behavior $\mathcal{B}_{k}$ is defined as in~\eqref{eq:behavior}, i.e.,
\begin{equation}
    \mathcal{B}_k = (w_k, \Theta_k, v_k, \Phi_k, G_k) \quad k = 1,\dots, M.
\end{equation}
\end{defn}

Here, note that the feasible sets $\Theta$ and $\Phi$, and the graph $G$ are different for different behaviors, that is, in switching between different behaviors the communication graphs of the robots may be time-varying. Moreover, based on Definition \ref{def:behavior} it is important to note the difference between {\it  behavior} and {\it controller}. The controller \eqref{eq:agent_controller} executed by the robots for a given behavior is obtained by selecting a proper pair of parameters $(\theta,\phi)$ from the sets $\Theta$ and $\Phi$. Indeed, consider a behavior $\mathcal{B}$ and let $x_t = [x_{1,t}^T, \dots, x_{N,t}^T]^T \in \mathbb{R}^{2N}$ be the ensemble states of the robots at time $t$. In addition, let $u_\mathcal{B}(x_t,\theta,\phi)$, where   $u_\mathcal{B} = [u_{1}^T, \dots,  u_{N}^T]^T \in \mathbb{R}^{2N}$, be the controllers of the robots  defined in \eqref{eq:agent_controller} for a feasible pair of parameters $(\theta,\phi)$. The ensemble dynamic of the robots associated $\mathcal{B}$ is then given as
\begin{equation} \label{eq:system_dynamics}
    \dot{x}_t =  u_\mathcal{B}(x_t,\theta,\phi).
\end{equation}
To further illustrate the difference between a behavior and its associated controller, we consider the following example about a formation control problem.
\begin{ex} \label{ex:formation}
Consider the formation control problem over a network of $4$ robots moving in a plane in Fig. \ref{fig:shapeExamples}, where the desired inter-robot distances are given by a vector $\theta=\{ \theta_{1},\dots,\theta_{5}\}$, with $\theta_{i}\in \mathbb{R}_+$. Here, agent $1$ acts as a leader and moves toward the goal $\phi \in \mathbb{R}^2$ (red dot). Note that the desire formation also implies the interaction structure between the robots (graph $G$ in our setting). The goal of the robots is to maintain their desired formation while moving to the red dot. To achieve this goal, one possible choice of the  edge-weight function of the controller \eqref{eq:agent_controller} is 
    \begin{equation} \label{eq:edgeFun_ex}
        w= \| x_i-x_j \| - \theta_k,   \quad \forall\; e_k=(i,j)\in E,
    \end{equation}
while the state-feedback term $v=0$ except for the one at the leader given as 
\begin{equation}
        v_1 = \phi - x_1.
\end{equation}
In this example, $\Phi$ is simply a subset of $\mathbb{R}^{2}$ while $\Theta$ is a set of geometrically feasible distances. Thus, given the formation control behavior $\mathcal{B}=(w, \Theta, v, \Phi, G)$, the controllers $u_\mathcal{B}(x,\theta,\phi)$ of the robots can be easily derived from \eqref{eq:agent_controller}.
    \begin{figure} \centering        \includegraphics[width=0.8\columnwidth]{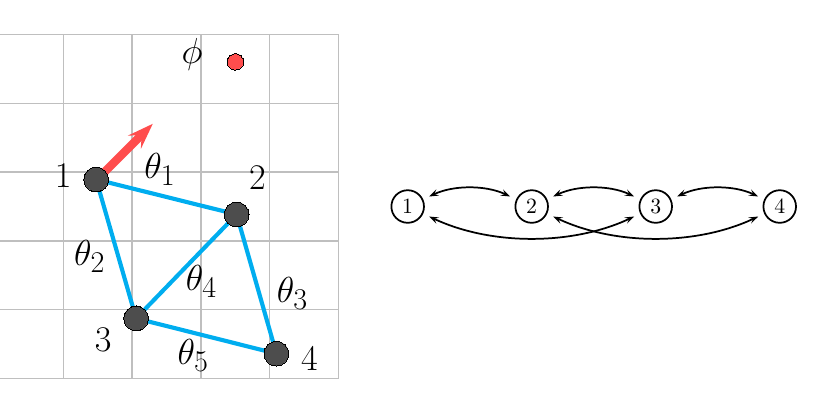}
        \caption{Example formation for a team of $4$ robots and one leader with goal $\phi$ (left). Interaction graph $G$ needed for the correct formation assembling (right).\label{fig:shapeExamples}}
    \end{figure}
\end{ex}
   
We  conclude this section with some comments on the formation control problem described above, which are the motivation for our study in the next section. In Example \ref{ex:formation}, one can choose a single behavior $\mathcal{B}_{i} \in \mathcal{L}_b$ together with a pair of parameters $(\theta,\phi)$ for solving the formation control problem \cite{mesbahi2010graph}.  This controller, however, is designed under the assumption that the
environment is static and known, i.e., the target $\phi$ in Example \ref{ex:formation} is fixed and known by the robots. Such an assumption is less practical since in many
applications, the robots are often operating in dynamically
evolving and potentially unknown environments; for example, $\phi$ is time-varying and unknown. On the other hand, while the formation control problem can be solved by using a single behavior, many practical complex tasks require the robots to implement more than one behavior \cite{nagavalli2017automated,pierpaoli2019sequential}. Our interest, therefore, is to consider the problem of selecting  a sequence of the behaviors in $\mathcal{L}_b$ for solving a given task while assuming that the state of the environment is unknown and possibly time-varying. In our setting, although the dynamic of the environment is unknown, we will assume that the robots can observe the state of the environment at any time through their course of action. We will refer to this setting as a behavior selection problem, which is formally presented in the next section. Finally, depending on the application, the environment can represent different quantities, (e.g. the target (red dot) in Example \ref{ex:formation}).

\section{Optimal Behavior Selection Problems}\label{sec:problemForm}
In this section, we present the problem of optimal behaviors selection over a network of robots, motivated by the reinforcement learning literature. In particular, consider a team of $N$ robots cooperatively operating in an unknown environment and their goal is to complete a given mission in a time interval $[0,t_{f}]$. Let $x_{t}$ and $e_{t}$ be the states of robots and environment at time $t\in[0,t_{f}]$, respectively. At any time $t$, the robots first observe the state of the environment $e_{t}$, select a behavior $\mathcal{B}_{t}$ chosen from the library $\mathcal{L}_{b}$, compute the pair of parameters $(\theta_{t},\phi_{t})$ associated with $\mathcal{B}_{t}$, and implement the resulting controller $u_{\mathcal{B}}(x_{t},\theta_{t},\phi_{t})$. The environment then moves to a new state $e_{t}'$ and the robots get a reward returned by the environment based on the selected behavior and tuning parameters. We assume that the rewards encode the given mission, which is motivated by the usual consideration in the literature of reinforcement learning \cite{Sutton2018_book}. That is, solving the task is equivalent to maximizing the total accumulated rewards received by the robots. In Section \ref{sec:applications}, we provide a few examples of how to design reward functions for particular applications. It is worth to point out that designing such a reward function is itself challenging and requires a good knowledge on the underlying problem \cite{Sutton2018_book}.        

One can try to solve the optimal behavior selection problem by using the existing techniques in reinforcement learning. However, this problem is in general intractable since the dimension of state space is infinite, i.e., $x_{t}$ and $e_{t}$ are continuous variables. Moreover, due to the physical constraint of the robots, it is infeasible for the robots to switch to a new behavior at every time instant. That is, the robots require a finite amount of time to implement the controller of the selected behavior. Thus, to circumvent these issues we next consider an alternate version of this problem.


Inspired by the work in~\cite{mehta2006optimal}  we introduce an interrupt condition $\xi: \mathcal{E} \mapsto \{0,1\}$, where $\mathcal{E}$ is given in \eqref{eq:energy}, defined as
\begin{equation}
    \xi(\mathcal{E}_t) = 1 \quad \text{if} \quad \mathcal{E}_t \leq \varepsilon
\end{equation}
and $\xi(\mathcal{E}_t)=0$ otherwise. Here $\varepsilon$ is a small positive threshold. In other words, $\xi(\mathcal{E}_t)$ represents a binary trigger with value $1$ whenever the network energy for a certain behavior at time $t$ is smaller than a threshold, that is, the current controller is nearly complete.  Thus, it is reasonable to enforce that the robots should not switch to a new behavior at time $t$ unless $\xi(\mathcal{E}_t) = 1$ for a given $\epsilon$. Based on this observation, given a desired threshold $\epsilon$, let $\tau_{i}$ be the switching time associated with the current behavior $\mathcal{B}$  defined as
\begin{equation} \label{eq:switchtime}
    \tau_i(\mathcal{B},\epsilon,t_{0}) = \min \{ t \geq t_0 \, | \, \mathcal{E}_t \leq \varepsilon \}.
\end{equation}
Consequently, the mission time interval $[0,t_f]$ is partitioned into $K$ switching times $\tau_{0},\ldots,\tau_{K}$ satisfying 
\begin{equation}
    0 = \tau_{0} \leq \tau_{1}\leq \ldots\leq \tau_{K} = t_{f},\label{def:switching}
\end{equation}
where each switching time is define as in~\eqref{eq:switchtime}. Note that the number of switching time $K$ depends on the accuracy $\epsilon$ and it is not known in advance. In this paper, we do not consider the problem of optimizing the number of switching times given a threshold $\epsilon$.      

At each switching time $\tau_{i}$, the robots choose one behavior $\mathcal{B}_{i}\in\mathcal{L}_{b}$ based on their current states $x_{\tau_{i}}$ and the environment state $e_{\tau_{i}}$. Next, they find a pair of parameters $(\theta_i,\phi_i)$ and implement the underlying controller $u_{\mathcal{B}_i}(x_{t},\theta_i,\phi_i)$ for $t\in[\tau_{i},\tau_{i+1})$. Based on the selected behaviors and parameters, the robots receive an instantaneous reward $\mathcal{R}(x_{\tau_{i}},e_{\tau_{i}},\mathcal{B}_{i},\theta_i,\phi_i)$ returned by the environment as an estimate for their selection. Finally, we assume that the states of the environment at the switching times belong to a finite set $\mathcal{S}$, i.e., $e_{\tau_{i}}\in\mathcal{S}$ for all $i$.     

Let $J$ be the accumulative reward received by the robots at the switching times in $[0,t_{f}]$
\begin{equation}
J = \sum_{i=0}^{K} \mathcal{R}(x_{\tau_{i}},e_{\tau_{i}},\mathcal{B}_{i},\theta_{\tau_{i}},\phi_{\tau_{i}}).
\end{equation} 
As mentioned above, the optimal behavior selection is equivalent to the problem of seeking a sequence of behaviors $\{\mathcal{B}_{i}\}$ from $\mathcal{L}_b$ at $\{\tau_{i}\}$ and the associated parameters $\{(\theta_i,\phi_i)\}\in \{\Theta_i\times\Phi_i\}$ so that the accumulative reward $J$ is maximized. This optimization problem can be formulated as follows
\begin{align}
\begin{aligned}
\underset{\mathcal{B}_i, \theta_i,\phi_i}{\text{maximize}} \quad &\sum_{i=0}^{K} \mathcal{R}(x_{\tau_{i}},e_{\tau_{i}},\mathcal{B}_{i},\theta_i,\phi_i)\\
\text{such that}\quad 
&\mathcal{B}_i \in \mathcal{L}_b,\;(\theta_i,\phi_i) \in \Theta_i\times \Phi_i\\
&e_{t+1} = f_e(x_{t}, e_{t})\\
&\dot{x} =  \,u_{\mathcal{B}_i}(x_{t},\theta_i,\phi_i),\; t\in[\tau_{i},\tau_{i+1}).
\end{aligned}\label{prob:opt_control}
\end{align}
where $f_e: \mathbb{R}^{2N} \times \mathbb{R}^{2} \mapsto \mathbb{R}^{2}$ is the unknown dynamic of the environment. Since $f_e$ is unknown, one cannot use dynamic programming to solve this problem. Thus, in the next section we propose a novel method for solving  \eqref{prob:opt_control}, which is a combination of $Q$-learning and online gradient descent. Moreover, by introducing the switching times $\tau_{i}$ computing the optimal sequence of behaviors using Q-learning is now tractable.     

\section{Q-Learning Approach For Behavior Selection} \label{sec:qlearning}
In this section we propose a novel reinforcement learning based method for solving problem \eqref{prob:opt_control}. Our method is composed of $Q$-learning and online gradient descent methods to find an optimal sequence of behaviors $\{\mathcal{B}_{i}^*\}$ and the associated parameters $\{(\theta^*_i,\phi^*_i)\}$, respectively. In particular, we maintain a Q-table, whose $(i,j)$ entry is the state-behavior value estimating the performance of behavior $\mathcal{B}_{j}\in\mathcal{L}_{b}$ given the environment state $j\in\mathcal{S}$. Thus, one can view the $Q$-table as a matrix  $Q\in\mathbb{R}^{S\times M}$, where $S$ is the size of $\mathcal{S}$ and $M$ is the number of behaviors in $\mathcal{L}_{b}$. The entries of Q-table are updated by using Q-learning while the controller parameters are updated by using the continuous-time online gradient method. These updates are formally presented in Algorithm \ref{alg:qlearning}.


In our algorithm, at each switching time $\tau_{i}$ the robots first observes the environment state $e_{\tau_i} = s\in \mathcal{S}$, and then select a behavior $\mathcal{B}_{m}$ with respect to the maximum entry in the $s$-th row of the $Q$-table with tie broken arbitrarily. Next, the robots implement the distributed controller $f_{\mathcal{B}_m}$ and use online gradient descent to find the best parameters $(\theta_{m},\phi_{m})$ associated with $\mathcal{B}_m$. Here the function $C_{t}$ represents the cost of implementing the controller at time $t$, which can be chosen priorily by the robots (e.g., $\mathcal{E}$ in \eqref{eq:energy}) or randomly returned by the environment. Based on the selected behavior and the associated controller, the robots receive an instantaneous reward $r_{i}$ while the environment moves to a new state $s'\in\mathcal{S}$. Finally, the robots updates the $(s,m)$ entry of the $Q$-table using the update of $Q$-learning method. It is worth to note that the Q-learning step is done in a centralized manner (either by the robots or a supervisory coordinator) since it depends on the state of the environment. Similarly, depending on the structure of the cost functions $C_t$ the online gradient descent updates can be implemented either distributedly or in a centralized manner. Finally, we note that the decentralized nature proper of each controller \eqref{eq:agent_controller} is preserved. 

\begin{algorithm}[h]
\caption{Q-learning algorithm for optimal behavior selection and tuning. The notation $\sim \mathcal{U}(\mathcal{O})$ is used to represent variables uniformly selected from a set $\mathcal{O}$} \label{alg:qlearning}
$x_{0} \sim \mathcal{U} (\mathbb{R}^{2N})$\;
$\mathcal{L}_b = \{\mathcal{B}_1,\dots,\mathcal{B}_M \}$\;
$Q(s,m) \sim \mathcal{U}(\mathbb{R}) \quad \forall m = 1,\ldots,M \, s\in \mathcal{S}$\;
\For{$m \in 1,\dots,M$}{
$\theta_{m} \sim \mathcal{U}(\Theta_m), \phi_{m} \sim \mathcal{U}(\Phi_m)$\;
}
$s\in\mathcal{S} \gets \text{Observe} \, e_{\tau_{i}} $ \;
\For{$i=0$ \KwTo $K$}{
    Select $m = \underset{\ell = 1,\ldots,M}{\arg\max}\;Q(s,\ell)$\;
    $(\bar{\theta}_t, \bar{\phi}_t) \gets (\theta_{m},\phi_{m})$ \;
    \While{$\mathcal{E}\geq \varepsilon$}{
        $\dot{x} = f_{\mathcal{B}_m}(x_t,\bar{\theta}_t,\bar{\phi}_t)$ \;
        $\dot{\bar{\theta}} = -\nabla_{\theta} \, C_t(x_{t},e_{\tau_i},\bar{\theta}_t,\bar{\phi}_t) \label{alg:btheta}$\; 
        $\dot{\bar{\phi}} = -\nabla_{\phi} \, C_t(x_{t},e_{\tau_i},\bar{\theta}_t,\bar{\phi}_t)\label{alg:bphi}$ \;
    }
    $(\theta_{m},\phi_{m}) \gets (\bar{\theta}_{t}, \bar{\phi}_{t})$\;
    $r_{i} \gets \text{from environment}$ \;
    $s'\in\mathcal{S} \gets \text{Observe} \, e_{\tau_{i+1}}$ \;
    $Q(s,m)=Q(s,m) + \epsilon\Big(r_i + \underset{j}{\max} \, Q(s',j) - Q(s,m)\Big)\label{alg:Qupdate}$\;
    $s \gets s'$
}
\end{algorithm}

\section{Applications} \label{sec:applications}

In this section we describe two implementations of our behaviors selection technique. For both examples, we considered a team of $5$ robots and a library of $5$ behaviors given as\footnote{These functions represent possible  behaviors of the robots. In particular, for the consistency with Example~\ref{ex:formation}, all individual agents terms are described by proportional controllers with unitary gains. Unlike Example~\ref{ex:formation}, here we let $\theta$ be a scaling factor, which are assumed to be fixed by the formation.} 
\begin{itemize}[leftmargin = 5.1mm]
\item [1)] Static formation:
\begin{equation*}
	u_i = \sum_{j \in \mathcal{N}_i} (\|x_i-x_j\|^2-(\theta\,\delta_{ij})^2)(x_j - x_i),
\end{equation*}
where $\delta_{ij}$ is the desired separation between robots $i$ and $j$, while $\theta\in\mathbb{R}$ is a shape scaling factor. 

\item [2)] Formation with leader:
\begin{align*}
	u_i &= \sum_{j \in \mathcal{N}_i} (\|x_i-x_j\|^2-\theta_{ij}^2)(x_j - x_i)\\
	u_l &= \sum_{j \in \mathcal{N}_i} \left( (\|x_i-x_j\|^2-\theta_{ij}^2)(x_i - x_i) \right) + (\phi-x_i), 
\end{align*}
where $\delta_{ij}$ and $\theta$ are defined as in the previous controller, while $\phi \in \Phi \subseteq \mathbb{R}^2$ is the leader's goal. Subscript $\ell$ leaders' controllers.

\item [3)] Cyclic pursuit:
\begin{equation*}
u_i = \sum_{j \in \mathcal{N}_i} R(\theta)\,(x_j - x_i) + (\phi-x_i),
\end{equation*}
where $\theta = 2 r\,\sin \frac{\pi}{N}$, $r$ is the radius of the cycle formed by the robots, and $R(\theta)\in SO(2)$. The point $\phi \in \Phi \subseteq \mathbb{R}^2$ is the center of the cycle.

\item [4)] Leader-follower:
\begin{align*}
	u_i &= \sum_{j \in \mathcal{N}_i} (\|x_i-x_j\|^2-\theta^2)(x_j - x_i) \\
	u_l &= \sum_{j \in \mathcal{N}_i} \left( (\|x_i-x_j\|^2-\theta^2)(x_j - x_i) \right) + (\phi-x_i),
\end{align*}
where $\theta$ is the separation between the agents and $\phi \in \Phi \subseteq \mathbb{R}^2$ is the leader's goal.

\item [5)] Triangulation coverage:
\begin{equation}
u_i = \sum_{j \in \mathcal{N}_i} (\|x_i-x_j\|^2-\theta^2)(x_j - x_i),
\end{equation} 
where $\theta$ is the separation between the agents in the triangulation. For all the behaviors considered above, we assume the following parameter spaces $\Theta = [0.05,1.1]$ and $\Phi = [-1,1]$. 
\end{itemize}
In both examples, we construct the state-action value function by implementing our propose method, Algorithm \ref{alg:qlearning}, on the Robotarium simulator ~\cite{pickem2017robotarium}, which captures a number of the features of the actual robots, such as the full kinematic model of the vehicles and collision avoidance algorithm.
 
\subsection{Convoy Protection}
First, we consider a {\it convoy protection} problem, where a team of robots must surround a moving target and maintain a single robot-to-target distance equal to a constant $\Delta$ at all times. Although this problem can be solved by executing a single behavior (e.g. cyclic-pursuit), it allows us to compare the performance of our framework against an ideal solution. The position of the target is denoted with $z_t$ and it's described by the following dynamics
\begin{equation}
    \dot{z} = v_z + \sigma,
\end{equation}
where $v_z$ is a constant velocity and $\sigma$ is a zero-mean Gaussian disturbance. In this case, the state of the environment at time step $t$ is considered to be the separation between robots' centroid $\bar{x}=\frac{1}{N}\sum_{i=1}^N\, x_i$ and the target
\begin{equation} \label{eq:envDef}
    e_t = \| \bar{x}_t - z_t\|,
\end{equation}
where $\|\cdot\|$ denotes the Euclidian norm. The reward provided by the environment at time $t$ is
\begin{equation}
    r_t = - \| e_t - \bar{x}_t \|^2  - \frac{1}{N}\sum_1^N (\|x_{i,t}-e_t \| - \Delta)^2,
\end{equation}
where the first term represents the proximity between centroid and target, while the second term weights the individual robot-to-target distance. Training is executed over $1000$ episodes, with an exponentially decaying $\varepsilon$-greedy policy.

\begin{figure}[h]
	\begin{center}
	\includegraphics[ width=\columnwidth ]{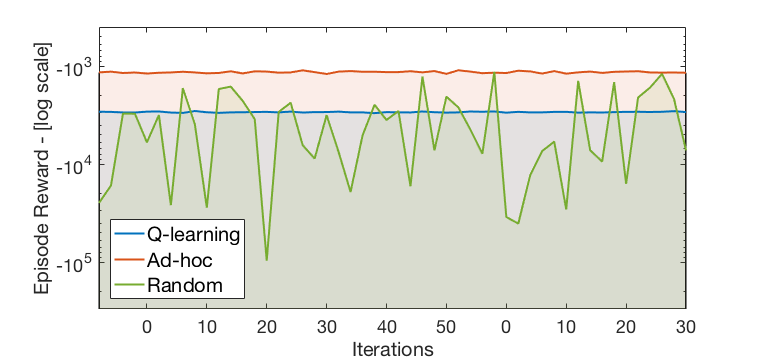}
		\caption{Comparison between accrued reward for $50$ different episodes of the convoy protection example. Rewards from trained model (blue) are compared with ad-hoc solution (red) and random behaviors/parameters selection (green). \label{fig:randomPi}}
	\end{center}
\end{figure}

The plot in Fig.~\ref{fig:randomPi} shows the collected rewards over a trial of $50$ episodes. The results from the trained model (blue) are compared against an ad-hoc ideal solution (red), where {\sc cyclic-pursuit} behavior is recursively executed with parameters $\theta$ and $\phi$ being selected so that the resulting cycle has radius $\Delta$ and is centered on the target's position. Finally, we show the rewards collected when the behaviors and parameters are selected uniformly at random (green). The variance of the results from random selection could be a direct indicator for the complexity of the problem. 

\vspace{-0.4cm}

\subsection{Simplified Object Manipulation}
\begin{figure*}[h]
	\begin{center}
		\includegraphics[trim={1cm 0.5cm 1cm 0cm}, width=0.65\columnwidth]{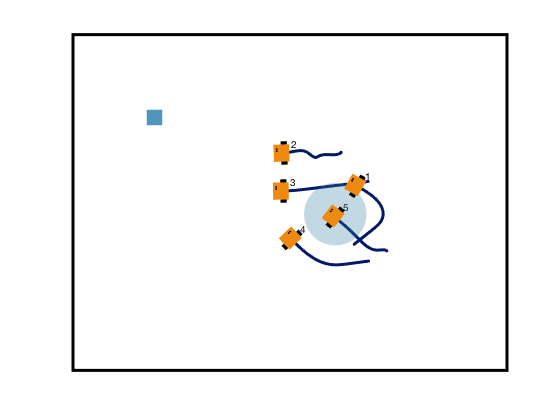}~
		\includegraphics[trim={1cm 0.5cm 1cm 0cm},width=0.65\columnwidth]{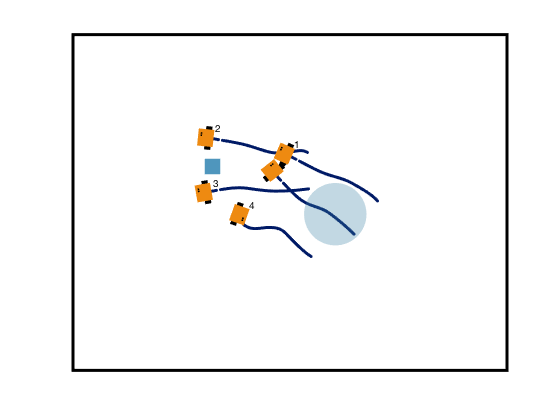}~
		\includegraphics[trim={1cm 0.5cm 1cm 0cm},width=0.65\columnwidth]{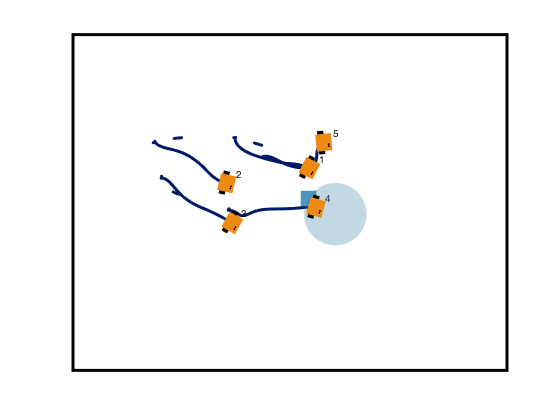}
		\caption{Screen shoot from Robotarium simulator during execution of the simplified object manipulation scenario at three different times. Executed behaviors are {\sc leader-follower} (left), {\sc Formation with leader}(center), and {\sc Cyclic-pursuit}(right). Blue square represents the object which is collectively transported towards the origin (blue circle). \label{fig:box_simu}}
	\end{center}
\end{figure*}

In the second example in Fig.~\ref{fig:box_simu}, we consider a team of robots tasked with moving an object from two points. Let $e_t$ represent the position of the object at time $t$. In order not to complicate the focus of the experiment, we assume a simplified manipulation dynamics. In particular, the box maintains its position if the closest robot is further than a certain threshold (i.e., object is not detected by the robots), otherwise it moves as $\bar{x}$. This manipulation dynamic guarantees that the task cannot be solved with fixed behavior and parameters since it is unknown to the robots. In this context, the robots get the following reward
\begin{equation}
    r_t = - (\kappa + \| e_t - \bar{e} \|),
\end{equation}
where $\kappa$ is a constant used to weight the running time until completion of an episode and $\| e_t - e_g \|$ is the distance between the box and its final destination $\bar{e}$.

\begin{figure}[h]
	\begin{center}
	\includegraphics[ width=\columnwidth ]{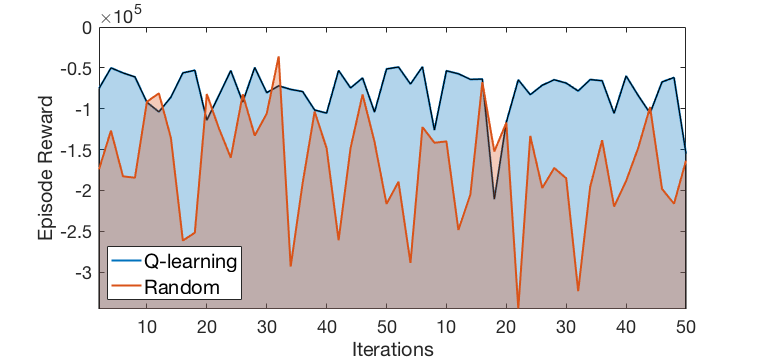}
		\caption{Comparison between accrued reward over $50$ episodes of the object manipulation example. Rewards from trained model (blue) are compared with random behaviors/parameters selection (red). \label{fig:box_results}}
	\end{center}
\end{figure}

\section{Conclusion} \label{sec:conc}
In this paper, we presented a reinforcement learning based approach for solving the optimal behavior selection problem, where the robots interact with an unknown environment. Given a finite library of behaviors, our technique exploits rewards collected through interaction with the environment to solve a given task that could not be solved by any single behavior. We also provide some numerical experiments on a network of robots to illustrate the effectiveness of our method. Future directions of this work include optimal design of behavior switching times and decentralized implementation of the Q-learning update. 
                                                   
\bibliographystyle{IEEEtran}
\bibliography{IEEEabrv,biblio.bib}


\end{document}